\documentclass{article}
\usepackage{spconf,amsmath,graphicx,hyperref}
\usepackage{amsmath,amsfonts}
\usepackage{algorithmic}
\usepackage{algorithm}
\usepackage{array}
\usepackage[caption=false,font=normalsize,labelfont=sf,textfont=sf]{subfig}
\usepackage{textcomp}
\usepackage{stfloats}
\usepackage{multirow}
\usepackage{url}
\usepackage{verbatim}
\usepackage{graphicx}
\usepackage{hyperref}
\usepackage{cite}
\usepackage{booktabs}
\usepackage{bbm}

\title{OilSAM2: Memory-Augmented SAM2 for Scalable SAR Oil Spill Detection}
\name{
  Shuaiyu Chen$^{1}$,
  Ming Yin$^{1}$,
  Peng Ren$^{2}$,
  Chunbo Luo$^{3}$,
  Zeyu Fu$^{1}$\sthanks{Corresponding Author.}
}
\address{
  $^{1}$ \normalsize Multimodal Intelligence Lab, Department of Computer Science, University of Exeter, United Kingdom \\
  \normalsize$^{2}$ College of Oceanography and Space Informatics, China University of Petroleum (East China), China.\\
  \normalsize$^{3}$ Department of Computer Science, University of Exeter, United Kingdom
}
\begin{document}
\ninept
\maketitle

\begin{abstract}

Segmenting oil spills from Synthetic Aperture Radar (SAR) imagery remains challenging due to severe appearance variability, scale heterogeneity, and the absence of temporal continuity in real-world monitoring scenarios. While foundation models such as Segment Anything (SAM) enable prompt-driven segmentation, existing SAM-based approaches operate on single images and cannot effectively reuse information across scenes. Memory-augmented variants (e.g., SAM2) further assume temporal coherence, making them prone to semantic drift when applied to unordered SAR image collections.
We propose OilSAM2, a memory-augmented segmentation framework tailored for unordered SAR oil spill monitoring. OilSAM2 introduces a hierarchical feature-aware multi-scale memory bank that explicitly models texture-, structure-, and semantic-level representations, enabling robust cross-image information reuse. To mitigate memory drift, we further propose a structure–semantic consistent memory update strategy that selectively refreshes memory based on semantic discrepancy and structural variation.
% To address these problems, we propose OilSAM2, a memory-augmented SAM2 framework designed for SAR oil spill segmentation. OilSAM2 introduces a multi-scale memory bank that models texture-level, structure-level, and semantic-level representations, which supports robust segmentation of both fragmented oil slicks and large continuous spill regions. A structure–semantic consistent memory update strategy is also introduced to control memory refreshing based on reliable structural and semantic cues. This design preserves boundary details and reduces semantic drift across images under noisy and heterogeneous SAR imaging conditions.
Experiments on two public SAR oil spill datasets demonstrate that OilSAM2 achieves state-of-the-art segmentation performance, delivering stable and accurate results under noisy SAR monitoring scenarios.  The source code is available at \url{https://github.com/Chenshuaiyu1120/OILSAM2}.

\end{abstract}

\section{Introduction}
Marine oil spills caused by offshore drilling accidents~\cite{bentz1976oil}, pipeline failures~\cite{doerffer2013oil}, and tanker incidents~\cite{alruzouq2020sensors} pose serious threats to marine ecosystems and coastal regions, so reliable identification of oil-contaminated areas is important for environmental monitoring and emergency response~\cite{solberg2007oil}. Synthetic aperture radar (SAR) has become a primary data source for oil spill monitoring because it operates independently of weather conditions and daylight~\cite{brekke2005oil}. With the increasing availability of large-scale SAR Earth observation data, deep learning-based methods have emerged as an effective solution for automatic oil spill detection~\cite{HTSM,brekke2005oil,mados}.

Convolutional neural network (CNN)-based approaches~\cite{satyanarayana2023oil,ronneberger2015u,li2023ds,chai2025transoilseg,chen2025enhancing} have demonstrated superior performance and robustness in SAR-based oil spill detection, outperforming traditional thresholding and handcrafted feature-based methods~\cite{fiscella2000oil,guo2014oil,temitope2020advances,li2023ds,dehghani2023oil} by learning discriminative representations directly from data. In addition,
% Encoder–decoder architectures are especially common, since hierarchical feature extraction and multi-scale decoding allow effective modeling of high-level semantic information and improve segmentation precision.
Transformer-based models~\cite{transformer1,transformer2} use self-attention to model relationships among all image patches at the same time, so global contextual information can be captured directly for segmentation~\cite{chai2025transoilseg}. However, self-attention introduces high computational cost, which limits the use of transformers in large-scale SAR oil spill monitoring~\cite{chai2025transoilseg,transformer1}. 
To reduce this burden, recent work has explored sequence modeling approaches based on state-space models (SSMs), which support long-range dependency modeling with linear computational complexity. Based on this idea, OSDMamba~\cite{OSDMamba} proposes a Mamba-based architecture for SAR oil spill segmentation. By using selective state-space scanning, OSDMamba captures long-range dependencies and global context while keeping computation efficient. 
% This design improves sensitivity to small and fragmented oil spill regions under strong class imbalance, but it does not include an explicit mechanism to maintain robustness and consistency across images collected under diverse environmental and imaging conditions.

% Recently, foundation models such as the Segment Anything Model (SAM)~\cite{kirillov2023segment} have attracted attention because of their strong generalization ability and prompt-driven flexibility across different visual domains. By enabling user-guided mask prediction with sparse prompts, SAM provides a unified segmentation framework that can be adapted to new tasks with limited task-specific training. Several studies have extended SAM to remote sensing and SAR oil spill detection, including SAM-Oil~\cite{wu2024compositional}, which integrates domain-specific prompt fusion strategies.

% In operational settings, SAR images are usually analyzed as unordered collections rather than temporally continuous sequences, so shared information across images can be useful for improving robustness. Memory-aware segmentation frameworks, such as SAM2~\cite{sam2} and Med-SAM2 \cite{medsam2}, offer a possible solution, but they are mainly designed for scenarios with temporal continuity or dense corrective signals. When applied directly to SAR oil spill monitoring, simple memory update strategies may introduce noise or cause semantic drift under diverse environmental and imaging conditions.
Recently, foundation models such as the Segment Anything Model (SAM)~\cite{kirillov2023segment}
have demonstrated the potential of adapting SAM through domain-specific prompt fusion \cite{wu2024compositional}.
However, SAM-based approaches operate on single images and cannot reuse information across scenes, which is often desirable in operational settings where SAR data are processed as collections rather than isolated samples. To address this, SAM2~\cite{sam2} introduces a memory mechanism for cross-image information reuse, though it has not yet been explored for SAR oil spill detection.

Adapting memory-based SAM frameworks to oil spill monitoring poses two main challenges. 1) Oil spills exhibit heterogeneous cues across feature scales, from fine-grained texture and speckle to higher-level structural and semantic representations, which a single-scale memory cannot capture effectively. 2) SAR oil spill images are unordered and lack temporal continuity, making memory retrieval sensitive to scene-specific variations such as sea state, backscatter intensity, and speckle noise, which can lead to the propagation of irrelevant information. These challenges motivate a structured and task-adaptive memory design for unordered SAR oil spill segmentation.

\begin{figure*}[t]
    \centering
    \includegraphics[width=\textwidth]{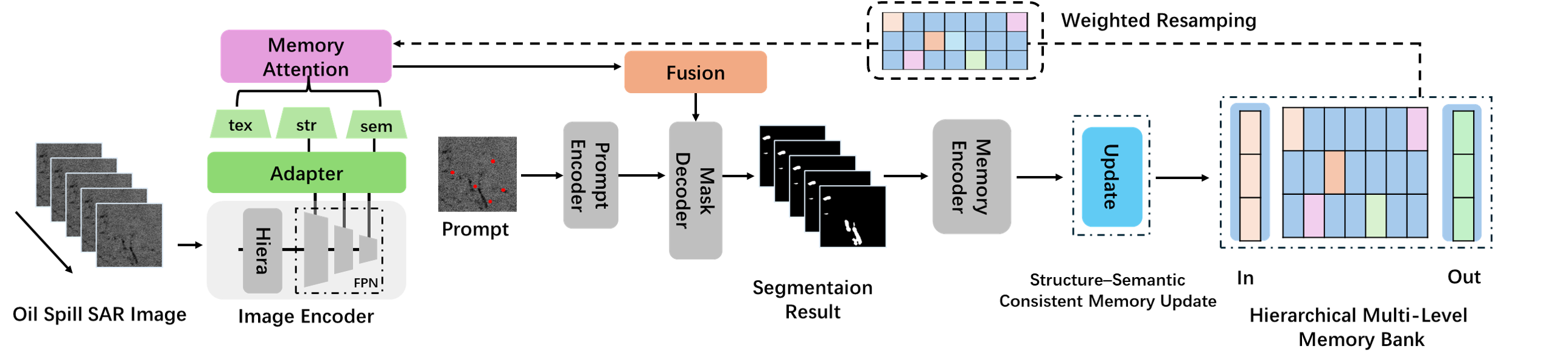}
    \vspace{-2em}
    \caption{\small Overview of the proposed OilSAM2 framework. Given an input SAR image and user prompt, hierarchical features are extracted and organized into texture-, structure-, and semantic-level representations. Each level interacts with a corresponding memory group via scale-wise attention. The retrieved features are adaptively fused and decoded to produce the segmentation mask, while a structure–semantic consistent update strategy regulates memory refreshing for robust reuse across unordered SAR images.}
    \vspace{-1em}
    \label{fig:m4dvis}
\end{figure*}

% In this paper, we propose OilSAM2, a SAM2-based architecture tailored for SAR oil spill segmentation, as illustrated in Fig.~\ref{fig:m4dvis}. OilSAM2 extends a memory-based segmentation approach \cite{medsam2} by introducing a structured multi-scale memory design that explicitly models texture-, structure-, and semantic-level representations. To address the unordered nature of SAR imagery and mitigate memory drift, we further adopt a structure–semantic consistent memory update strategy, which preserves reliable cross-image information while maintaining compatibility with the original SAM interaction workflow. Experimental results on two publicly available SAR oil spill datasets demonstrate that OilSAM2 achieves state-of-the-art performance and exhibits improved robustness under diverse imaging and environmental conditions.
In this paper, we propose OilSAM2, a SAM2-based architecture tailored for SAR oil spill segmentation, as illustrated in Fig.~\ref{fig:m4dvis}. OilSAM2 is built on a memory-based SAM framework \cite{medsam2} by introducing a hierarchical feature-aware memory bank that explicitly models texture-, structure-, and semantic-level representations. To accommodate the unordered nature of SAR imagery and mitigate semantic drift, we further propose a structure–semantic consistent memory update strategy that regulates memory refreshing using reliability cues. This design enables robust cross-image information reuse while preserving compatibility with the original SAM interaction paradigm. Extensive experiments on two publicly available SAR oil spill datasets demonstrate that OilSAM2 achieves strong segmentation performance and improved robustness under diverse imaging and environmental conditions.

\section{Method}
\label{sec:method}

\subsection{Problem Formulation}
We study prompt-driven SAR oil spill segmentation over an unordered image
collection $\mathcal{X}=\{x_t\}_{t=1}^{T}$, where the index $t$ denotes the
processing order rather than chronological time. Given an input SAR image
$x_t \in \mathbb{R}^{H\times W}$ and a user prompt $p$ (e.g., clicks or bounding
boxes), the goal is to predict a pixel-wise segmentation mask.
We denote the number of semantic classes by $C$ (with $C{=}2$ for binary oil-spill
segmentation). The ground-truth label map is
$y_t \in \{0,1,\ldots,C{-}1\}^{H\times W}$, and the model produces a soft
prediction $\hat{Y}_t \in [0,1]^{C \times H \times W}$, where
$\sum_{c=0}^{C-1}\hat{Y}_t(c,x,y)=1$ for each pixel $(x,y)$. The final hard mask is
$\hat{y}_t(x,y)=\arg\max_c \hat{Y}_t(c,x,y)$.

Following the general formulation in \cite{medsam2}, each image $x_t$ is encoded by an
image encoder $E_{\mathrm{img}}$ to obtain visual features
$F_t = E_{\mathrm{img}}(x_t)$, while the prompt $p$ is mapped to a prompt embedding
$Q = E_{\mathrm{prompt}}(p)$ via a prompt encoder $E_{\mathrm{prompt}}$.
A memory bank $\mathcal{M}_{t-1}=\{(K_j,V_j)\}$ maintains a compact set of key--value
representations derived from previously processed images, where $K_j$ denotes a
memory key and $V_j$ is the associated value used for segmentation guidance.

Given the current features $F_t$, a subset of memory entries
$\tilde{\mathcal{M}}_{t-1} \subseteq \mathcal{M}_{t-1}$ is retrieved based on
feature similarity and fused with the image features and prompt embedding through
a memory attention module $A(\cdot)$. The aggregated representation is decoded by a
mask decoder $D(\cdot)$ to yield the per-pixel class probabilities:
\begin{equation}
\hat{Y}_t = \mathrm{Softmax}\!\left(D\!\left(A\!\left(F_t, \tilde{\mathcal{M}}_{t-1}, Q\right)\right)\right).
\end{equation}
After prediction, the model encodes the output (e.g., $\hat{Y}_t$ or $\hat{y}_t$
together with $F_t$) via a memory encoder to form a new memory entry, which is
inserted into the memory bank. The memory bank is periodically updated to retain
a compact and representative set of entries.

While this formulation provides an effective baseline for SAR oil spill segmentation,
it does not explicitly account for the unordered nature of SAR imagery or the
multi-level feature heterogeneity of oil spills. These limitations motivate the
structured memory design and update strategies introduced in the following sections.
% it implicitly assumes that all stored memory entries are equally reliable and transferable across scenes. In contrast to Medical SAM2, our problem setting does not involve temporally ordered inputs. We focus on segmentation of unordered single-frame SAR images, and no intrinsic temporal relationship exists between samples. Although the index $t$ is kept for notational consistency with the SAM2 formulation, it does not indicate a temporal order in this setting. Instead, $t$ indexes independent image–memory interaction steps, and each image $x_t$ is treated as an independent observation.In practice, 
% SAR oil spills exhibit heterogeneous characteristics across spatial scales:
% fine-grained texture and speckle patterns support local discrimination,
% mid-level structural cues capture elongated slick morphology, and higher-level
% semantic representations help suppress visually similar look-alike phenomena.
% A single-scale memory representation is therefore insufficient to model these
% diverse properties.

% These observations motivate the need for a more selective and task-adaptive memory
% modeling strategy, which we address in the following section.
% large variations in sea state, backscatter intensity, and speckle noise can bias similarity-based retrieval toward scene-specific artifacts, causing irrelevant or low-quality memory entries to be propagated and ultimately degrading segmentation robustness.
\subsection{Hierarchical Feature-Aware Memory Bank}
SAR oil spill appearance varies across feature abstraction levels. Fine-grained
texture and speckle statistics are essential for detecting fragmented slicks,
mid-level structural patterns capture elongated morphology, and higher-level
semantic representations help suppress visually similar look-alike phenomena.
Encoding these heterogeneous cues within a single memory space can blur
level-specific information and degrade retrieval reliability.
As illustrated in Fig.~\ref{fig:m4dvis}, our framework exploits the hierarchical representations
produced by modern image encoders to construct a structured multi-level memory bank. Modern image encoders~\cite{sam2,medsam2} naturally produce hierarchical feature
representations at different depths. Given an input $x_t$, we denote the resulting
features as $\{F_t^{(s)}\}_{s\in\{\text{tex},\text{str},\text{sem}\}}$, corresponding
to early-, middle-, and late-stage encoder outputs. Leveraging this hierarchy, we
construct a structured multi-level memory bank that explicitly stores complementary
information at three feature levels: texture (\texttt{tex}), structure
(\texttt{str}), and semantic (\texttt{sem}). The memory at step $t{-}1$ is defined as
\begin{equation}
\mathcal{M}_{t-1}
=\{\mathcal{M}_{t-1}^{\text{tex}},
  \mathcal{M}_{t-1}^{\text{str}},
  \mathcal{M}_{t-1}^{\text{sem}}\},
\end{equation}
where each group contains key--value pairs
\begin{equation}
\mathcal{M}_{t-1}^{(s)} =
\{(K_j^{(s)}, V_j^{(s)})\}_{j=1}^{N_s},
\quad s \in \{\text{tex}, \text{str}, \text{sem}\}.
\end{equation}
Each feature level interacts exclusively with its corresponding memory group,
preserving level-specific representations and improving robustness under diverse
SAR imaging conditions.

\subsection{Scale-Adaptive Memory Fusion}

Although the multi-level memory bank supports structured information reuse,
features retrieved from different levels are heterogeneous and may contribute
unequally across oil spill morphologies and sea states.  
% For example, texture-level cues are often critical for fragmented slicks, whereas structure- and semantic-level cues tend to dominate for large or elongated spill regions. 
We therefore introduce a scale-adaptive fusion module to integrate complementary information across feature levels.

Memory retrieval is performed independently at each level. For
$s \in \{\text{tex}, \text{str}, \text{sem}\}$, the feature $F_t^{(s)}$ is first
mapped into a shared embedding space via a lightweight adapter, followed by
scale-wise memory attention to obtain
\begin{equation}
\tilde{F}_t^{(s)} =
A\!\left(F_t^{(s)},\, \bar{\mathcal{M}}_{t-1}^{(s)},\, Q_t\right),
\end{equation}
where $\bar{\mathcal{M}}_{t-1}^{(s)}$ denotes the retrieved memory entries at level
$s$.

We then align $\{\tilde{F}_t^{(s)}\}$ to the same spatial resolution and compute a level-wise response
score
\begin{equation}
e_t^{(s)}=\|\tilde{F}_t^{(s)}\|_1,
\end{equation}
which serves as a lightweight indicator of the contribution of level $s$ for the current input.
The scores are normalized to obtain fusion weights
\begin{equation}
\gamma_t^{(s)}=
\frac{\exp(e_t^{(s)})}{\sum_{s'\in\{\text{tex},\text{str},\text{sem}\}}\exp(e_t^{(s')})}.
\end{equation}
Finally, the fused representation is computed as
\begin{equation}
F_t^*=\sum_{s\in\{\text{tex},\text{str},\text{sem}\}}\gamma_t^{(s)}\,\tilde{F}_t^{(s)},
\end{equation}
which is forwarded to the mask decoder for final segmentation prediction.

\subsection{Structure-Semantic Consistent Memory Update}
Sections 2.2 and 2.3 describe how OilSAM2 retrieves and fuses
multi-level memory for mask prediction.
We now introduce a structure-semantic consistent memory update strategy. Unlike temporally coherent settings~\cite{medsam2}, our SAR inputs
are unordered; thus, updating memory after every prediction can accumulate scene-specific artifacts (e.g., sea state and speckle), which are later retrieved and cause semantic drift.

To regulate memory refreshing, we maintain lightweight structure/semantic prototypes,
$F_{\text{mem}}^{\text{str}}$ and $z_{\text{mem}}^{\text{sem}}$, summarizing the current memory
state. For the current sample, we extract a semantic descriptor
\begin{equation}
z_t^{\text{sem}}=\mathrm{GAP}\!\left(F_t^{\text{sem}}\right),
\end{equation}
and measure semantic discrepancy via cosine distance
\begin{equation}
\Delta_{\text{sem}}=1-\cos\!\left(z_t^{\text{sem}},\,z_{\text{mem}}^{\text{sem}}\right).
\end{equation}
At the structure level, we measure boundary variation using gradient magnitude
\begin{equation}
\Delta_{\text{str}}=\frac{1}{HW}\left\|\nabla F_t^{\text{str}}-\nabla F_{\text{mem}}^{\text{str}}\right\|_1 .
\end{equation}
where $H$ and $W$ denote the spatial height and width of the feature maps.

We update semantic memory when $\Delta_{\text{sem}}>\tau_{\text{sem}}$ and structure memory when
$\Delta_{\text{str}}>\tau_{\text{str}}$; the texture group is refreshed only together with a
semantic or structural update to preserve cross-level consistency. When triggered, new memory
content $M_t^{(s)}$ is encoded from the current sample and integrated via exponential moving average:
\begin{equation}
M_{\text{mem}}^{(s)}\leftarrow (1-\alpha)\,M_{\text{mem}}^{(s)}+\alpha\,M_t^{(s)},
\end{equation}
where $\alpha$ controls the update rate, limiting abrupt shifts under noisy SAR conditions.

\begin{table*}[t]\small
\setlength{\tabcolsep}{3pt}
\centering\caption{\small Comparison of OilSAM2 with other oil spill segmentation methods on the M4D Dataset~\cite{M4D}.\label{tab:table1}}
\centering
\begin{tabular}{c c c c c c c}
\toprule
Model & SeaSurface(\%, \(\uparrow\)) & Oil Spill(\%, \(\uparrow\)) & Look-alike(\%, \(\uparrow\)) & Ship(\%, \(\uparrow\)) & Land(\%, \(\uparrow\)) & mIoU(\%, \(\uparrow\))\\
\midrule

Unet~\cite{ronneberger2015unet} & 93.90 & 53.79 & 39.55 & 44.93 & 92.68 & 64.97 \\

LinkNet~\cite{Linknet} & 94.99 & 51.53 & 43.24 & 40.23 & 93.97 & 64.79 \\

PSPNet~\cite{PSPNet} & 92.78 & 40.10 & 33.79 & 24.42 & 86.90 & 55.60 \\

Deeplabv2~\cite{Deeplabv2} & 94.09 & 25.57 & 40.30 & 11.41 & 74.99 & 49.27 \\

Deeplabv2(msc)~\cite{Deeplabv2} & 95.39 & 49.28 & 31.26 & 88.65 & 93.97 & 62.83 \\

Deeplabv3+~\cite{dpv3p} & 96.43 & 53.38 & 55.40 & 27.63 & 92.44 & 65.06 \\
% 96.05	51.60	55.60	52.55	91.81	69.52
TransOilSeg\cite{chai2025transoilseg} & 97.02 & 61.38 & 62.41 & 33.49 & 94.39 & 69.74 \\
YOLOv8-SAM ~\cite{wu2024compositional} & 94.34 & 41.84 & 48.15 & 52.48 & 87.65 & 64.89 \\
SAM-OIL  ~\cite{wu2024compositional} & 96.05 & 51.60 & 55.60 & 52.55 & 91.81 & 69.52 \\
OSDMamba~\cite{OSDMamba} & 96.47 & 65.59 & 47.57 & 46.85 & 94.76 & 70.25 \\
OilSAM2( Ours) & 95.10 & \textbf{65.92} & 54.16 & 56.18 & 92.07 & \textbf{72.62} \\

\bottomrule
\end{tabular}
\end{table*}

\begin{table*}[t]
\small
\setlength{\tabcolsep}{4pt}
\centering
\caption{\small Comparison of OilSAM2 with other oil spill segmentation methods (on SOS Dataset~\cite{SOS}).}
\label{tab:table2}
\begin{tabular}{lcccccccc}
\toprule
\multirow{2}{*}{Model} & 
\multicolumn{2}{c}{mIoU (\%, $\uparrow$)} & 
\multicolumn{2}{c}{F1-score (\%, $\uparrow$)} & 
\multicolumn{2}{c}{Recall (\%, $\uparrow$)} & 
\multicolumn{2}{c}{Precision (\%, $\uparrow$)} \\
\cmidrule(lr){2-3} \cmidrule(lr){4-5} \cmidrule(lr){6-7} \cmidrule(lr){8-9}
 & PALSAR & Sentinel-1 & PALSAR & Sentinel-1 & PALSAR & Sentinel-1 & PALSAR & Sentinel-1 \\
\midrule
U-Net~\cite{ronneberger2015unet}     & 81.63 & 81.46 & 82.20 & 86.10 & 77.84 & 81.22 & 83.08 & 85.61 \\
D-LinkNet~\cite{Linknet}             & 81.58 & 82.32 & 84.57 & 87.08 & 85.48 & 85.22 & 83.69 & 85.22 \\
Deeplabv3~\cite{dpv3p}               & 81.79 & 82.94 & 84.26 & 87.70 & 84.31 & 84.76 & 83.02 & 88.08 \\
CBD-Net~\cite{SOS}                   & 83.31 & 83.42 & 84.84 & 87.87 & 86.75 & 87.32 & 84.20 & 91.20 \\
Medical SAM2~\cite{medsam2}                     & 82.14 & 80.64 & 83.04 & 82.13 & 83.77 & 85.92 & 81.07 & 85.64 \\
TransOilSeg                       & 83.08 & 82.27 & 96.73 & 93.18 & 96.52 & 93.07 & 96.94 & 93.43 \\
OilSAM2 (Ours)                       & \textbf{84.20} & \textbf{83.67} & 83.98 & 88.75 & 87.64 & 86.51 & 83.58 & 90.39 \\
\bottomrule
\end{tabular}
\end{table*}

\begin{table}[t]
\centering
\caption{\small Ablation study of components (SOS Dataset PALSAR).\\ \footnotesize{Scale adapters are replaced by linear projections in the ablation study.}}
\begin{tabular}{lcc}
\hline
Configuration & mIoU & Prec. \\
\hline
Baseline \cite{medsam2} & 82.14 & 81.07 \\

Baseline + Multi-Scale Fusion & 82.27 & 81.13 \\

Baseline + Structure–Semantic Update & 83.64 & 81.68 \\

Baseline + Multi-scale Memory Bank & 83.53 & 82.04 \\

Baseline + Fusion + Update & 83.92 & 82.16 \\
%最后一组_01 6
Baseline + Fusion + Memory Bank & 83.85 & 82.24 \\

Baseline + Update + Memory Bank & 84.10 & 83.06 \\

OILSAM2 (Full Model) & \textbf{84.20} & \textbf{83.58} \\
\hline
\end{tabular}
\label{tab:ablation}
\end{table}
\section{Experiments}
\label{sec:typestyle}
\subsection{Datasets}
We conduct experiments on two widely used SAR oil spill Datasets: the
M4D dataset~\cite{M4D} and the Deep-SAR Oil Spill (SOS) dataset the ~\cite{SOS}. 
The M4D dataset~\cite{M4D} consists of 1,002 training and 110 testing SAR images with a resolution of 1250×650 pixels, covering five semantic categories: sea surface, oil spill, look-alike phenomena, ships, and land.

The SOS dataset~\cite{SOS} contains 8,070 labeled SAR image patches (256×256 pixels) derived from ALOS PALSAR and Sentinel-1A imagery over the Gulf of Mexico and Persian Gulf. It is split into 6,455 training and 1,615 testing samples, covering diverse sea states and spill morphologies for robust oil spill detection.

\subsection{Implementation Details}
Our framework is implemented in PyTorch and trained on a single NVIDIA L40 GPU. The AdamW optimizer is adopted with an initial learning rate of 0.001, weight decay of 0.0001, and batch size of 4. The model is trained using a weighted binary cross-entropy loss, following the setting used in ~\cite{medsam2}. During training, the image encoder, prompt encoder, and the original memory attention and memory encoder modules of SAM2~\cite{sam2} are kept frozen. The proposed scale adapters, multi-scale memory components, and structure–semantic update modules are trained together with the SAM mask decoder.

% The model is trained for 100 epochs with early stopping. We fine-tune a pre-trained SAM2 model for SAR oil spill segmentation under a prompt-conditioned setting, and segmentation is guided by randomly sampled point prompts. During training, the image encoder, prompt encoder, memory attention module, and memory encoder are kept frozen, while only the SAM mask decoder is updated.

\subsection{Comparison with State-of-the-Art}
Table~\ref{tab:table1} presents quantitative results compared with classical CNN-based (U-Net~\cite{ronneberger2015unet}, PSPNet~\cite{PSPNet}, DeepLabv3+~\cite{dpv3p}), transformer-based, and SAM-derived methods. Our method achieves consistent improvements across both the M4D~\cite{M4D} and SOS datasets. On the M4D dataset (Table~\ref{tab:table1}), our approach obtains a 14.32\% increase in Oil Spill IoU and a 3.10\% increase in overall mIoU compared to SAM-Oil, demonstrating the effectiveness of memory-augmented segmentation in handling complex spill morphologies. Notably, OilSAM2 not only improves the detection of oil spills but also maintains competitive performance across other semantic categories such as look-alikes, ships, and land, indicating better generalization to diverse marine targets.

% On the SOS dataset (Table~\ref{tab:table2}), our framework also surpasses classical CNNs (U-Net, D-LinkNet), advanced segmentation networks (Deeplabv3, CBD-Net), and foundation-model-based approaches (SAM2). Specifically, OilSAM2 achieves the highest mIoU of 84.20\% on PALSAR and 83.67\% on Sentinel-1, alongside superior F1-score and recall. These gains confirm that the structure–semantic consistent update stabilizes boundary delineation, the multi-scale memory bank captures both fine-grained and large-scale slicks, and the one-prompt segmentation paradigm significantly reduces interaction costs while ensuring accuracy.
On the SOS dataset (Table~\ref{tab:table2}), the proposed framework outperforms classical CNN-based models such as U-Net and D-LinkNet, advanced segmentation networks including DeepLabv3 and CBD-Net, and foundation-model-based approaches such as SAM2. OilSAM2 achieves the highest mIoU of 84.20\% on PALSAR and 83.67\% on Sentinel-1, and it also shows a higher F1-score and recall. These results show that the structure–semantic consistent update helps stabilize boundary delineation, the multi-scale memory bank captures oil slicks at different scales, and the multi-scale fusion strategy supports adaptive aggregation of complementary representations.
\subsection{Ablation Study}

\begin{figure}[t]
    \centering
    \includegraphics[width=\linewidth]{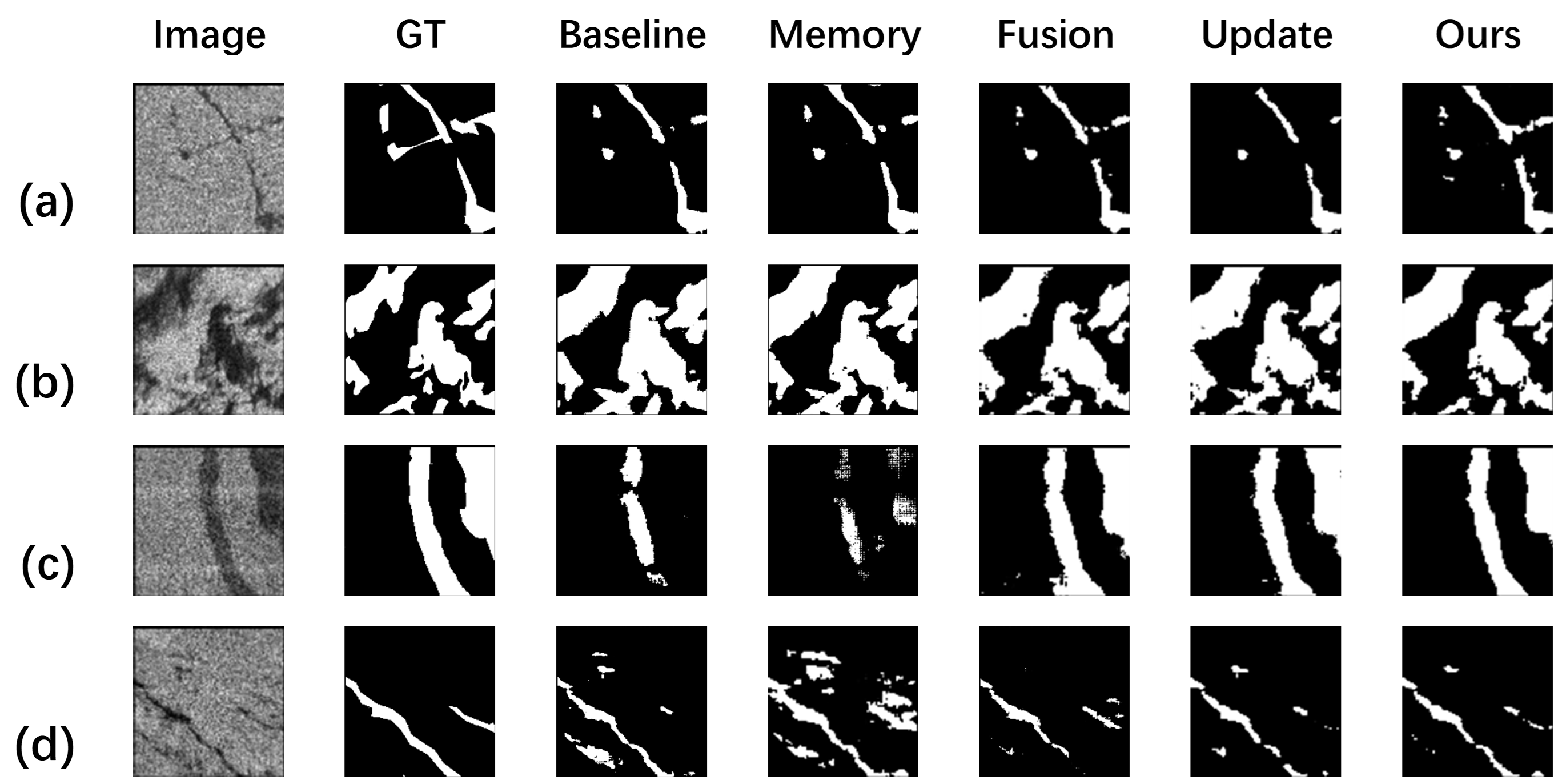}
    \vspace{-2em}
    \caption{\small Qualitative results of the ablation study(On SOS Dataset).}
    \label{fig:ablation}
\end{figure}

To evaluate the contribution of each proposed component, we conduct an ablation study on the SOS dataset. Quantitative results are reported in Table~\ref{tab:ablation}, and qualitative comparisons are shown in Figure~\ref{fig:ablation}.

Starting from the baseline model \cite{medsam2}, we examine each component separately. Adding the multi-scale fusion strategy alone results in a small improvement, which indicates that adaptive feature aggregation provides limited benefit when memory augmentation is not used. Adding the structure–semantic consistent update produces a larger gain in both mIoU and precision. This result suggests that updating memory based on semantic and structural cues helps stabilize memory evolution and preserve boundary information. Introducing the multi-scale memory bank alone also leads to a clear performance improvement, which confirms that organizing memory across texture, structure, and semantic levels is effective.

We then study the combined effects of different components. As shown in Table~\ref{tab:ablation}, combining the structure–semantic update with the multi-scale memory bank yields the largest gain among all two-component variants. This result reflects a strong interaction between structured memory representation and controlled memory updating. Combining the fusion strategy with either component also improves performance, but the gains are smaller.

When all components are included, the full OilSAM2 model achieves the best performance across all metrics. The results indicate that the three components complement each other. Multi-scale fusion aggregates information across different representation levels, the multi-scale memory bank captures heterogeneous spatial patterns, and the structure–semantic update maintains stable memory evolution.

Qualitative results in Figure~\ref{fig:ablation} are consistent with the quantitative analysis. Compared with reduced variants, the full model produces clearer boundaries, improves detection of small and fragmented oil slicks, and reduces missed and false detections under complex sea conditions.

% \begin{table}[t]
% \centering
% \caption{\small Ablation study of proposed components on the SOS Dataset (PALSAR).}
% \begin{tabular}{lcc}
% \hline
% Configuration & mIoU & Prec. \\
% \hline
% Baseline (SAM2~\cite{sam2}) & 82.14 & 81.07 \\

% Structure–Semantic Update & 83.64 & 81.68 \\

% Multi-scale Memory Bank & 83.53 & 82.04 \\

% One-Prompt Segmentation (Full Model) & \textbf{84.20} & \textbf{83.58} \\
% \hline
% \end{tabular}
% \label{tab:ablation}
% \end{table}

\vspace{-1em}
% The ablation results in Table~\ref{tab:ablation} and Figure~\ref{fig:ablation} clearly demonstrate the contribution of each proposed component. Starting from the SAM2 baseline, which already achieves competitive performance, we observe that introducing the structure–semantic update yields a notable improvement in both mIoU and accuracy. This indicates that selectively refreshing the memory bank based on edge variations and semantic consistency effectively stabilizes memory evolution and preserves boundary details, which are often blurred in SAR oil spill segmentation. The addition of the multi-scale memory bank further enhances performance by capturing complementary information across shallow texture, intermediate structure, and deep semantics. This hierarchical representation improves robustness when detecting both small fragmented slicks and large continuous oil spill regions. Finally, integrating the one-prompt segmentation paradigm results in the full model, which achieves the highest scores. This demonstrates that propagating a single prompt through the memory bank not only reduces human interaction costs but also ensures consistent and accurate segmentation across large collections of unordered SAR images. 

% The qualitative results (red boxes in Figure~\ref{fig:ablation}) further confirm these findings, highlighting clearer boundary delineation, improved detection of small patches, and fewer missed or false detections compared with reduced variants of our model.

\section{Conclusion}

% In this paper, we propose OilSAM2, a memory-augmented framework for marine oil spill detection from SAR imagery. The framework targets challenges related to scale variation, semantic ambiguity, and robustness in large-scale monitoring settings. Experiments on benchmark SAR datasets show that OilSAM2 outperforms CNN-based, transformer-based, and SAM-derived baselines. The experimental results reflect three main aspects of the proposed approach. The structure–semantic consistent memory update strategy helps preserve boundary integrity and reduces semantic drift during memory evolution. The multi-scale memory bank stores texture-level, structure-level, and semantic-level representations, and this design supports reliable segmentation of both fragmented oil slicks and large continuous spill regions. OilSAM2 also introduces a memory-enhanced prompt-conditioned segmentation framework that accumulates cross-image information while keeping the original SAM interaction setting, which improves robustness and cross-image consistency. These design choices improve structural robustness, scale adaptability, and practical applicability for SAR-based marine oil spill monitoring. 
In this paper, we presented OilSAM2, a memory-augmented segmentation framework for unordered SAR oil spill monitoring. OilSAM2 extends SAM2 with a hierarchical feature-aware multi-scale memory bank and a structure–semantic consistent memory update strategy, enabling robust cross-image information reuse under diverse SAR imaging conditions. The multi-scale memory design captures complementary texture-, structure-, and semantic-level representations, supporting accurate segmentation of both fragmented and large-scale oil spills, while the proposed update mechanism prevents semantic drift by regulating memory refreshing based on reliability cues. Experimental results on benchmark SAR datasets demonstrate that OilSAM2 consistently outperforms CNN-based, transformer-based, and SAM-derived baselines, providing improved robustness and cross-image consistency for effective oil spill monitoring.
% In this paper, we proposed OilSAM2, a memory-augmented framework for marine oil spill detection from SAR imagery, tackling the challenges of scale variability, semantic ambiguity, and high interaction costs in large-scale monitoring. Experiments on benchmark SAR datasets demonstrated that OilSAM2 consistently outperforms CNN-based, transformer-based, and SAM-derived baselines. Both experiment results highlight three key innovations: a structure–semantic consistent update that preserves boundaries and prevents semantic drift, a multi-scale memory bank that robustly handles both fragmented and large-scale spills, and a one-prompt segmentation paradigm that reduces human effort while maintaining accuracy. These contributions enhance structural robustness, scale adaptability, and scalability for practical marine oil spill monitoring. Looking ahead, we plan to extend OilSAM2 to multi-modal data and SAR time series, adapt it to other pollution scenarios such as plastic debris and algal blooms, and explore lightweight designs for real-time operational deployment.

\bibliographystyle{IEEEbib}
\bibliography{refs}

\end{document}